\documentclass[10pt,twocolumn,letterpaper]{article}

\usepackage{iccv}              

\newcommand{\todo}[1]{{\color{red}#1}}

\definecolor{iccvblue}{rgb}{0.21,0.49,0.74}
\definecolor{cvprblue}{rgb}{0.21,0.49,0.74}
\usepackage[pagebackref,breaklinks,colorlinks,allcolors=iccvblue]{hyperref}
\usepackage{multirow,multicol,colortbl,makecell}
\usepackage[table]{xcolor}
\usepackage{tabularx}
\usepackage[para]{footmisc}

\def\paperID{8526} 
\def\confName{ICCV}
\def\confYear{2025}
\renewcommand{\todo}{\textcolor{black}}
\newcommand{\note}{\textcolor{black}}

\newcommand{\cb}{\textcolor{cvprblue}}
\title{Sim-DETR: Unlock DETR for Temporal Sentence Grounding}

\author{
  \textbf{Jiajin Tang}$^1$$^{*}$,
  \textbf{Zhengxuan Wei}$^1$$^{*}$,
  \textbf{Yuchen Zhu}$^1$,
  \textbf{Cheng Shi}$^1$,
  \textbf{Guanbin Li}$^2$,
  \textbf{Liang Lin}$^2$,
  \textbf{Sibei Yang}$^2$$^{\dagger}$\\
\textsuperscript{1}ShanghaiTech University \quad 
\textsuperscript{2}School of Computer Science and Engineering, Sun Yat-sen University \\
\texttt{\normalsize{\{tangjj, weizhx2022\}@shanghaitech.edu.cn}}\hspace{0.5cm}\texttt{\normalsize{yangsb3@mail.sysu.edu.cn}}\\
}
\begin{document}
\maketitle
\renewcommand{\thefootnote}{\fnsymbol{footnote}}
\footnotetext{
  $^{*}$ Equal contribution. \quad
  $^{\dagger}$ Corresponding author is Sibei Yang.
}
\begin{abstract}
Temporal sentence grounding aims to identify exact moments in a video that correspond to a given textual query, typically addressed with detection transformer (DETR) solutions. However, we find that typical strategies designed to enhance DETR do not improve, and may even degrade, its performance in this task. We systematically analyze and identify the root causes of this abnormal behavior: (1) conflicts between queries from similar target moments and (2) internal query conflicts due to the tension between global semantics and local localization. Building on these insights, we propose a simple yet powerful baseline, Sim-DETR, which extends the standard DETR with two minor modifications in the decoder layers: (1) constraining self-attention between queries based on their semantic and positional overlap and (2) adding query-to-frame alignment to bridge the global and local contexts. Experiments demonstrate that Sim-DETR unlocks the full potential of DETR for temporal sentence grounding, offering a strong baseline for future research. Code is available at \small \url{https://github.com/SooLab/Sim-DETR}.
\end{abstract}    
\vspace{-4mm}
\section{Introduction}
\label{sec:intro}

\begin{figure}
    \centering
    \includegraphics[width=0.91\linewidth]{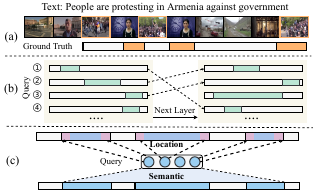}
    \vspace{-2.3mm}
    \caption{Illustration of (a) temporal sentence grounding task, (b) phenomenon of random matching across multiple layers, and (c) the challenge of achieving the balance between focus on boundary for span prediction and alignment with global semantics.}
    \label{fig:1}
    \vspace{-5.2mm}
\end{figure}

Video has become a dominant media on the internet, with short-form content rapidly expanding and achieving exponential growth in reach over recent years~\cite{han2020self,kim2021relational,huang2018makes,huang2023free38}.
Instead of passively watching entire videos, users now prefer to target specific segments of interest, with natural language descriptions serving as an intuitive and flexible way to convey intent~\cite{tang2023temporal71,li2023progressive,xie2023ra}. 
This amplifies interest in the research topic of temporal sentence grounding~\cite{yang2024taskweave,sun2024trdetr,liu2022umt,jang2023eatr,lin2023univtg,moon2023qd-detr,lei2021qvhighlights,moon2023cgdetr,lee2025bamdetr,zhang2020tan,lei2021detecting}, which aims to locate one or more semantically aligned segments within an untrimmed video according to a given natural language sentence, \note{as shown in Figure~\ref{fig:1}\cb{a}.} 

\begin{figure*}
    \centering
    \includegraphics[width=1\linewidth]{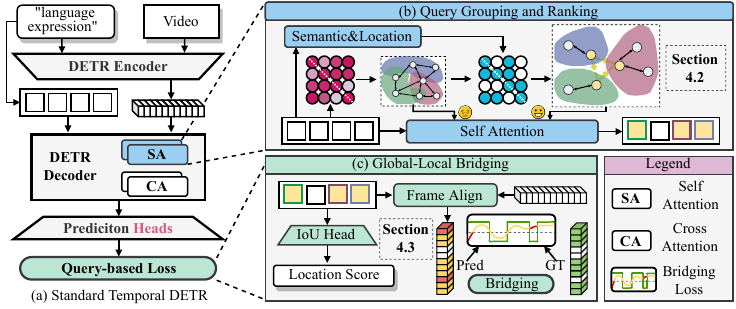}
    \vspace{-2mm}
    \caption{Framework of the proposed Sim-DETR, featuring two simple yet powerful modifications: Query Grouping and Ranking to mitigate cross-query conflicts and  Global-Local Bridging to address global-local conflicts within each query.}
    \label{fig:7}
    \vspace{-4mm}
\end{figure*}

Early methods either align sentences with predefined temporal proposals for selection~\cite{mao2020generation,yadav2020unsupervised} or directly predict moment spans through cross-modal interactions between language and frames~\cite{liu2022memory,liu2022reducing,yang2022entity}. 
Recent advances center on integrating the Detection Transformer (DETR)~\cite{carion2020detr} for object detection into the temporal sentence grounding, leveraging its query-based proposal detection framework to eliminate hand-crafted proposal components and deliver superior grounding performance with high efficiency~\cite{jang2023eatr,lee2025bamdetr,sun2024trdetr,moon2023cgdetr,moon2023qd-detr,lei2021detecting}. 
They leverage moment queries—often several times more numerous than ground truth segments—in the multi-layer decoder, which are initialized randomly~\cite{moon2023cgdetr,sun2024trdetr,lei2021detecting}, generated from sentences~\cite{moon2023qd-detr,jang2023eatr}, or extracted as event units from the video~\cite{lee2025bamdetr}, to locate and search the most suitable matches for target segments. 
A one-to-one label assignment is applied in the training objective to facilitate non-redundant predictions for each ground-truth segment. 
The overall architecture of the DETR-based temporal sentence grounding framework is shown in \note{Figure~\ref{fig:7}\cb{a}.}

However, we observe that strategies generally effective for DETR in other tasks, such as object detection, like increasing the number of queries or decoder layers, do not improve performance in temporal sentence grounding and may even degrade it \note{(Sec~\ref{sec:3.1})}.  

This motivates us to revisit the specific characteristics of temporal sentence grounding that give rise to the distinct challenges in applying DETR to this task. 
In a series of investigations on query-segment alignment \note{(Sec~\ref{sec:3.2})}, we verify that performance degradation primarily stems from 
the difficulty in identifying \textbf{\textit{distinct local boundaries}} for target segments that share \textbf{\textit{highly similar global semantics}}, \textbf{\textit{leading to inherent conflicts both between and within queries.}} 
By diagnosing the similarities between queries and between queries and frames, we uncover two conflicts that result in learning puzzles for queries: 
\begin{itemize}
\item \textit{Queries corresponding to distinct target segments exhibit high similarity, leading to random matching with various targets and creating learning puzzles for the queries.} The presence of multiple semantically similar target segments results in the high similarity between their corresponding queries, making them difficult to clearly distinguish \note{(Sec~\ref{sec:3.3})}. This causes each query to potentially match well with multiple segments, leading to significant randomness in the one-to-one matching for each target segment \note{(Figure~\ref{fig:1}\cb{b})}. We observe that across different decoder layers, the queries matched for optimization can vary considerably \note{(Sec~\ref{sec:3.2})}. 
\item \textit{Each query faces an inherent conflict between encoding the global semantics of its corresponding segment and decoding its local boundary.} To align well with a segment, a query must encode the segment's global semantics from the start to the end frame, such as the entire process of a person holding a cup. However, the query also needs to directly predict the segment's boundary, requiring focus on the local frames near the boundary (\note{Figure~\ref{fig:1}\cb{c}}). This creates a trade-off: the query either prioritizes global semantics or local boundaries, making it challenging to optimize both simultaneously \note{(Sec~\ref{sec:3.3})}.  
\end{itemize}
Building on these findings, we propose a simple yet strong baseline for video temporal grounding, Sim-DETR. The framework builds on the standard DETR-based temporal sentence grounding architecture, \textit{\textbf{with two minor yet straightforward design modifications to the decoder layers}} to address the two conflicts between and within queries (Figure~\ref{fig:7}), respectively. 
First, we adjust the self-attention weights between queries based on their pairwise correlation, encouraging indistinguishable queries to focus on different contexts, reducing their similarity, and allowing the most suitable query to gather information from related queries to refine its prediction ({Figure~\ref{fig:7}\cb{b}}). 
Second, we introduce a query-to-frame matching and loss term, accounting for the alignment between the query and each frame within the segment, as shown in \note{Figure~\ref{fig:7}\cb{c}}. This ensures that all frames, not just those near the boundary, contribute to segment localization. The full sequence of all target frames, from start to finish, acts as a bridge connecting the global semantics to the local boundary. 

Empirically, we indeed observe that our minor modifications (1) significantly reduce conflicts between queries corresponding to different target segments \note{Figure~\ref{fig:3})},  (2) improve the alignment between the query's global semantic attention to frames and its boundary prediction \note{(Figure~\ref{fig:5})}, and  (3) lead to more consistent query predictions across layers~\note{(Figure~\ref{fig:4})}. Furthermore, (4) increasing the number of queries and decoder layers no longer causes performance degradation~\note{(Figure~\ref{fig:2})}. Finally, due to the consistent matching and prediction of queries, (5) our Sim-DETR not only achieves substantial performance gains but also accelerates convergence \note{(as shown in Appendix)}. 
All the above observations and advantages, including the minor modifications, are intended to unlock the potential of query-based frameworks for temporal sentence grounding and provide a simple yet strong baseline for future research. 

\noindent Our contributions are summarized as follows: 
\begin{itemize}
    \item We systematically analyze the root causes of abnormal behavior in the DETR-based temporal sentence grounding framework, identifying two key conflicts: (1) target segments with highly similar semantics but distinct temporal localization create learning puzzles between queries, and (2) a trade-off between relying on global semantics for matching and local boundaries for localization within queries. 

    \item Based on our analysis, we propose two modifications to form Sim-DETR: (1) a simple adjustment to the self-attention to resolve conflicts between queries and (2) the introduction of a query-to-frame matching to reconcile global semantics and local localization within queries.
    \item Our Sim-DETR, a simple yet powerful baseline, achieves consistent and significant improvements across all benchmarks. More importantly, it eliminates observed anomalies and exhibits faster convergence. 
\end{itemize}

\section{Related Work}
\label{sec:related_work}
\noindent \textbf{Temporal Sentence Grounding (TSG)} aims to identify specific video moments that align with given language expressions. As computer vision advances across neural architectures, generation, detection, segmentation, and multimodal tasks~\cite{chen2024survey12,huang2025mvtokenflow39,shi2024plain66,yang2021bottom81,lin2021structured46,dai2024curriculum20,shi2024part2object64,zheng2023ddcot88,shi2023logoprompt63}, temporal understanding requires task-specific optimizations. Early methods fall into proposal-free and proposal-based categories. Proposal-free methods use end-to-end frameworks by directly regressing boundary coordinates~\cite{lu2019debug, chen2020gdp, zeng2020drn} or predicting frame-level boundary likelihood~\cite{ghosh2019excl, zhang2020vslnet}. Proposal-based approaches densely sample candidate segments via sliding windows~\cite{gao2017ctrl, anne2017mcn, liu2018role, zhang2019tcmn, ge2019mac, jiang2019slta}, temporal anchors~\cite{lin2023univtg, liu2024tuning, chen2018tgn, zhang2019man, yuan2019scdm, zhang2020tan}, or multimodal feature similarity~\cite{xu2019qspn, chen2019sap, xiao2021bpnet, liu2021apgn, liu2022slp}. Recently, DETR has shown exceptional effectiveness in detection tasks. \cite{lei2021qvhighlights,woo2022explore} pioneered DETR for TSG. Subsequent studies enhance cross-modal representations~\cite{moon2023qd-detr, sun2024trdetr, xiao2024uvcom, moon2023cgdetr}, improve decoder queries with semantic/spatial information~\cite{jang2023eatr, lee2025bamdetr, sun2024rgtr}, or explore joint training with other temporal tasks~\cite{sun2024trdetr,yang2024taskweave}.

\noindent \textbf{Detection Transformers (DETR)}~\cite{carion2020detr} is a pioneering end-to-end framework for object detection and segmentation that employs transformers to make direct set-based predictions~\cite{ease-detr,zhu2025rethinking,shi2023edadet62,shi2024plain66,shi2024part2object64,tang2023temporal71,tang2023contrastive70}. However, DETR has notable limitations, especially its high computational complexity and slow convergence, impacting its practical efficiency. To address these issues, several derivative models have emerged. Some methods~\cite{zhu2020deformable, yao2021efficient, wang2021pnp, roh2021sparse} improve computational efficiency by reducing redundant calculations within the DETR architecture. Other models~\cite{gao2021smca, sun2021tsp, dai2021dynamic} refine the attention mechanism for more efficient information processing and resource utilization. Additionally~\cite{meng2021conditional, wang2022anchor, liu2022dab} enhance convergence by embedding spatial information directly into the query design, which accelerates object localization. Recent models~\cite{li2022dn,zhang2022dino} tackle optimization challenges in DETR's bipartite matching process by integrating denoising strategies during training, improving both convergence and performance. 
Despite recent advancements in enhancing DETR for object detection, there is a lack of systematic studies addressing the unique challenges of applying DETR to TSG and the potential difficulties in query optimization.

\section{Probing DETR's Inherent Behavior for TSG}
\label{sec:3}

This study begins with the observation that enhancements effective for object detection in DETR~\cite{carion2020detr} do not apply to temporal sentence grounding (TSG), as discussed in Sec~\ref{sec:3.1}. 
To explore the reasons behind DETR's failure to enhance, this section presents preliminary studies to reveal the behavior of queries in the DETR decoder, as queries are central to predicting proposals. Specifically, we examine the similarity between queries and outputs in Sec~\ref{sec:3.2}, and the attention between queries and inputs (\ie, frames) in Sec~\ref{sec:3.3}.  
These help reveal the underlying cause, and our modification addressing it not only eliminates the abnormal phenomena but also has a positive side effect (Sec~\ref{subsec:effect}). 

\noindent\textbf{Experimental Settings.} Unless otherwise specified, our default experimental setup involves conducting statistical analyses on the full validation set of QVHighlights~\cite{lei2021qvhighlights}. CG-DETR~\cite{moon2023cgdetr} and TaskWeave~\cite{yang2024taskweave}, which serve as baselines for DETR-based TSG due to their strong performance while retaining the core vanilla DETR mechanism, are used with their standard configurations. 

\subsection{Abnormal Phenomena}
\label{sec:3.1}

\textbf{Motivation.} DETR is originally proposed for object detection, and increasing the number of queries and decoder layers typically improves performance. We explore whether this trend also applies to TSG. 

\noindent\textbf{Setting.}
\textcolor{black}{Both CG-DETR and TaskWeave default to 10 queries, with 3 and 2 decoder layers, respectively. This experiment specifically increases the number of queries and decoder layers to isolate their effects, keeping all other hyperparameters and training protocols constant.}

\begin{figure}[h]
    \centering
    \vspace{-2pt}
    \includegraphics[width=0.96\linewidth]{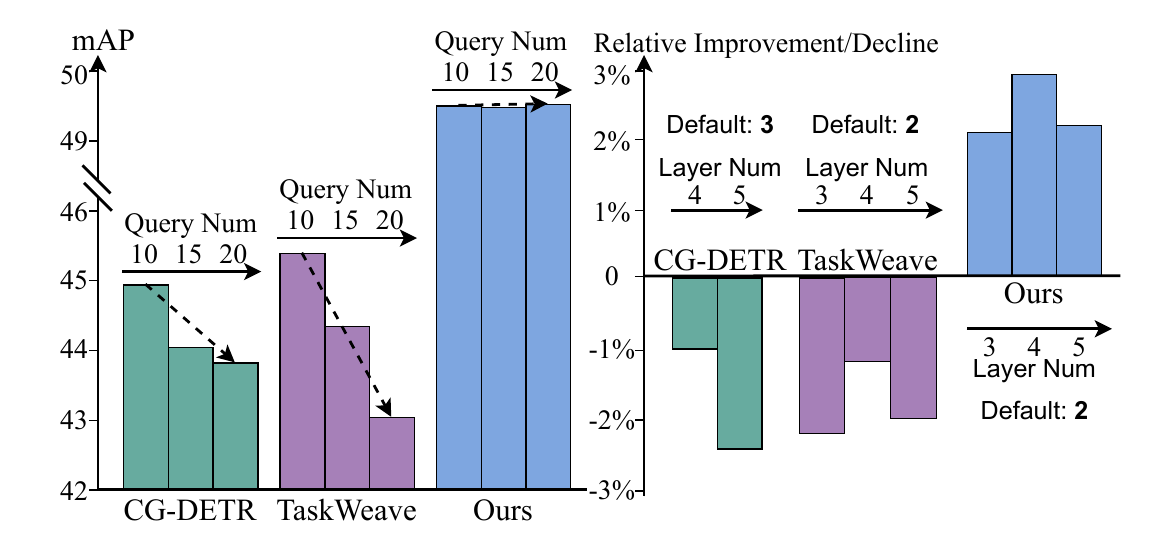}
    \vspace{-8pt}
    \caption{\textcolor{black}{The impact of the number of queries and decoder layers.}}
    \label{fig:2}
    \vspace{-3mm}
\end{figure}

\noindent\textbf{Result \& Discussion.} As shown in Figure~\ref{fig:2}, increasing the number of queries to 15 and 20 led to progressively larger performance declines, with a drop of over -2\% with TaskWeave~\cite{yang2024taskweave} at 20 queries. Similarly, increasing the number of decoder layers caused a drop from -1\% to -2.5\%. In contrast, our Sim-DETR mitigates this phenomenon and achieves a slight improvement, demonstrating its robustness. This observation motivates an in-depth exploration of decoder layer, particularly the role of queries (see Sec~\ref{sec:3.2}).

\subsection{Similar Target Segments Cause Query Conflicts}
\label{sec:3.2}

\textbf{Motivation.} Initially, we hypothesize that the lack of performance improvement with increased queries is due to query redundancy. However, performance declined, suggesting that conflicts between queries, rather than redundancy, prevent compatibility with existing queries. Similarly, conflicts between queries across layers may explain the ineffectiveness of increasing the number of layers. This motivates us to explore (1) the relationships between queries to identify their potential conflicts and (2) their variations across layers to reveal the impact at different layers. 

\noindent\textbf{Setting.}
\textcolor{black}{(1) To evaluate query relationships, we assess similarities between queries corresponding to the same segment (intra-segment similarity) and those from different segments (inter-segment similarity). Specifically, we identify the corresponding segment for each query based on the maximum IoU between the query's prediction and all ground-truth (GT) segments during inference. Queries whose IoU with the corresponding segment is 0.5 or greater are included in our analysis to ensure meaningful corresponding. We then compute feature similarities between queries associated with the same segment (intra) and those matched to different segments (inter). 
(2) To analyze variations across layers, we introduce the metric of cross-layer matching consistency. Specifically, during training optimization, each GT segment is uniquely matched to a query through bipartite matching at each decoding layer. We assess match consistency by measuring the proportion of segments that retain their corresponding queries across two consecutive decoder layers.}

\vspace{-2pt}
\begin{figure}[h]
    \centering
    \vspace{-8pt}
    \includegraphics[width=0.93\linewidth]{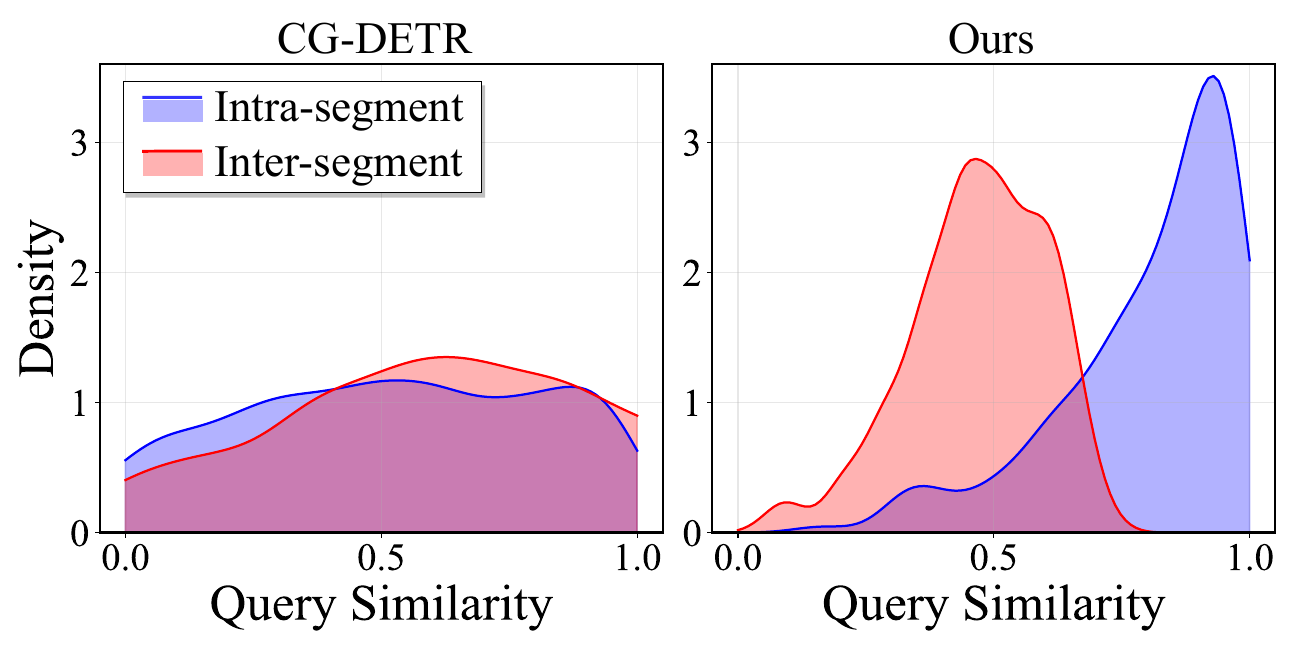}
    \vspace{-8pt}
    \caption{\textcolor{black}{Distribution of query similarity for intra-segment (blue) and inter-segment (red) pairs. Our method can effectively distinguish intra-segment and inter-segment queries, ensuring more stable query-segment associations and reducing conflicts in query assignments across different segments.}}
    \label{fig:3}

    \centering
    \includegraphics[width=0.9\linewidth]{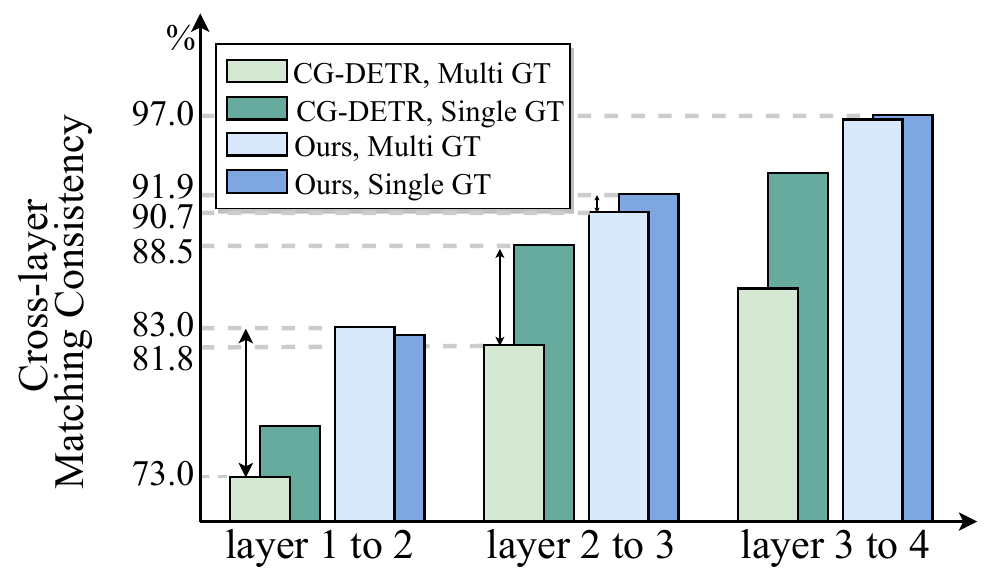}
    \vspace{-8pt}
    \caption{\textcolor{black}{Measuring cross-layer matching consistency (y-axis, defined in Line-260) of segments across two consecutive decoding layers (x-axis). Our method achieves higher consistency.}}
    \label{fig:4}
   \vspace{-3mm}
\end{figure}

\noindent\textbf{Result \& Discussion.} (1) As shown in Figure~\ref{fig:3}, CG-DETR fails to distinguish query similarities between inter- or intra-segment groups, indicating that queries are not clearly grouped with their corresponding segments. \textbf{\textit{Consequently, the same query oscillates between segments, lacking stable and consistent corresponding.}} Figure~\ref{fig:4} further supports our hypothesis, demonstrating that cross-layer matching consistency in CG-DETR is lower than ours. (2) We attribute this to multiple GT segments in a video sharing the same linguistic semantics. \textbf{\textit{The high semantic similarity between GT segments results in high similarity between their corresponding queries.}} Figure~\ref{fig:4} supports this, showing that the consistency for cases with multiple GT segments is lower than for those with a single segment. 

\subsection{Global Matching vs. Local Localization}
\label{sec:3.3}
\textbf{Motivation.} We further question whether the inconsistency in matching is solely due to query conflicts from multiple GT segments. If so, why does the previous method still show roughly 25\% mismatched queries in cases with a single target segment? Therefore, we shift our focus from conflicts between queries to conflicts within individual queries. A query serves dual roles: aligning with global linguistic semantics (global matching) and precisely localizing the segment, particularly its boundaries (local localization). This prompts us to explore whether conflicts arise from these roles within a single query. 

\begin{figure}[tb]
    \centering
    \includegraphics[width=1\linewidth]{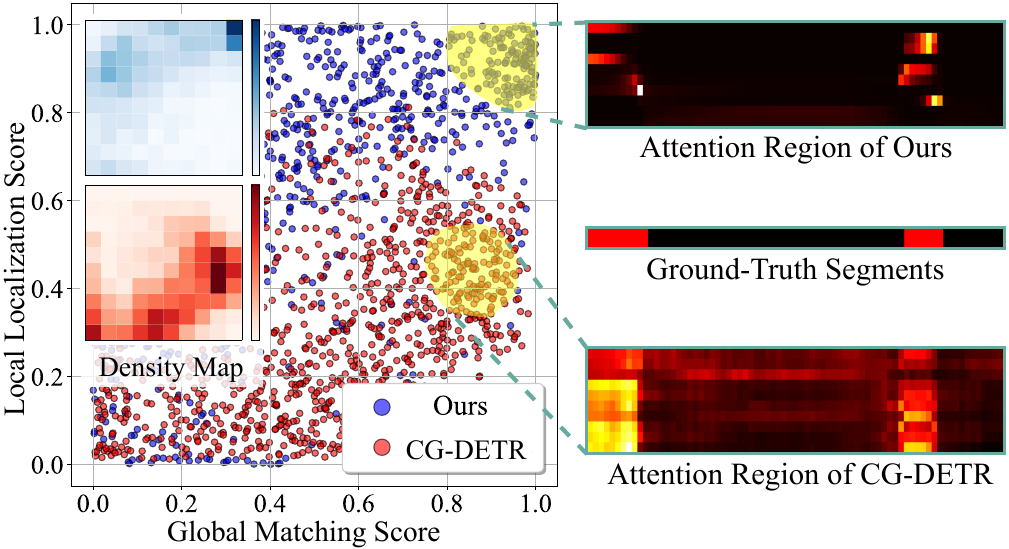}
    \caption{\textcolor{black}{Global matching score (defined in Line-290) vs. local localization score (defined in Line-293). 
    Compared to CG-DETR, our method concentrates a query's attention significantly more within a single GT segment, rather than dispersing it across multiple segments, leading to improved local localization scores.} 
    }
    \label{fig:5}
      \vspace{-5mm}
\end{figure}

\noindent\textbf{Setting.} \textcolor{black}{(1) To evaluate global matching, we use the query's classification confidence score, as it reflects the confidence of the query being referenced by the global semantics of the linguistic description. (2) To evaluate local localization, we extract attention scores between each query and frame from the cross-attention module in the last decoder layer, as they indicate query-based frame localization. We then compute the IoU score between this attention and the query’s corresponding GT segment, reflecting the accuracy of the query’s localization at the frame level.} 

\noindent\textbf{Result \& Discussion.} \textcolor{black}{As shown in Figure~\ref{fig:5}, CG-DETR (marked in red) may yield low local localization scores despite high global matching scores, indicating that \textbf{\textit{while queries align well with linguistic descriptions, they fail to achieve precise localization.}} Further analysis reveals that query attention may span frames in multiple GT segments \textcolor{black}{(see Figure~\ref{fig:5} ``Attention")}, leading to a low local score for the query's specific GT segment it corresponds to.}

\subsection{Positive Side Effect}
\label{subsec:effect}
\textbf{Convert Abnormal to Normal:} Our Sim-DETR effectively (1) distinguishes queries within the same target segment and across different segments in Figure~\ref{fig:3}, (2) achieves consistent matching across decoder layers in Figure~\ref{fig:4}, and (3) aligns with global semantics while ensuring precise localization in Figure~\ref{fig:5}.

\noindent\textbf{Positive Side Effect:} (1) More importantly, Sim-DETR maintains stable and robust performance with increased queries and decoder layers (see Figure~\ref{fig:2}). (2) By eliminating the abnormal phenomena, it also achieves significantly faster convergence compared to previous methods, as shown in Appendix~\textcolor{cvprblue}{C}.

\section{Method}
\label{sec:4}
We present a simple yet effective TSG baseline that resolves query conflicts in existing DETR-based TSG frameworks~\cite{lei2021qvhighlights, moon2023qd-detr, jang2023eatr, moon2023cgdetr, sun2024trdetr, xiao2024uvcom, yang2024taskweave, lee2025bamdetr} through two minor yet key modifications. First, we introduce the standard DETR-based~\cite{carion2020detr} TSG architecture (Sec~\ref{sec:4.1}). 
\note{Sec~\ref{sec:4.2} introduces our first modification, which adjusts the self-attention mechanism between queries in the decoder to avoid conflicts among them, based on the observations presented in Sec~\ref{sec:3.2}. Next, Sec~\ref{sec:4.3} presents another improvement that addresses the conflict between global matching and local localization within the query (discussed in Sec~\ref{sec:3.3}) by introducing query-to-frame matching loss, which serves as the bridge between the global and local perspectives.}

\subsection{Standard DETR-based TSG Architecture}
\label{sec:4.1}
In TSG, the goal is to identify video segments with temporal spans that correspond to a given language expression. 

\noindent\textbf{Feature Extraction.} Previous DETR-based works~\cite{moon2023qd-detr, sun2024trdetr, yang2024taskweave, lee2025bamdetr} employ CLIP~\cite{radford2021clip} and SlowFast~\cite{feichtenhofer2019slowfast} to extract video features, represented as \todo{$\mathcal{T} \in \mathbb{R}^{N \times C}$}, where $N$ and $C$ stands for the number of frames and the hidden dimension, respectively.
For the language expression, typically a sentence with \( L \) tokens, the CLIP text encoder is used to obtain token-level features, resulting in $\mathcal{W} \in \mathbb{R}^{L \times C}$. 

\noindent\textbf{Multimodal Encoder.} After feature extraction, DETR-based TSG methods~\cite{moon2023qd-detr, jang2023eatr, moon2023cgdetr, sun2024trdetr} employ a multimodal encoder to fuse these video and language features, producing multimodal video features, denoted as \todo{$\hat{\mathcal{T}}\in \mathbb{R}^{N\times C}$}. The key component of the multimodal encoder is a cross-attention mechanism~\cite{vaswani2017transformer}, where video features \(\mathcal{T}\) act as queries, while word features \(\mathcal{W}\) serve as keys and values. 

\noindent \textbf{Temporal Sentence Decoding.} In the decoder, the model initializes a set of query to gather global information from the video, which is then decoded into local spans in \todo{the span prediction head.} Specifically, given initialized queries \todo{$\mathcal{Q} \in \mathbb{R}^{M \times C}$}, where $M$ is the query number, and the multimodal video features $\hat{\mathcal{T}}$, the decoder facilitates interaction between queries and gathers global information through interleaved self-attention and cross-attention modules, as follows: 

\[
\begin{aligned}
     &
    \begin{cases}
        \mathcal{Q}_q^{\text{SA}} = \operatorname{FC}_q^{\text{SA}}(\mathcal{Q}),  \mathcal{Q}_k^{\text{SA}} = \operatorname{FC}_k^{\text{SA}}(\mathcal{Q}),  \mathcal{Q}_v^{\text{SA}} = \operatorname{FC}_v^{\text{SA}}(\mathcal{Q}), \\ 
        \mathcal{Q}^{\text{SA}} = \operatorname{FC}_o^{\text{SA}}\left(\operatorname{softmax}\left(\frac{\mathcal{Q}_q^{\text{SA}}(\mathcal{Q}_k^{\text{SA}})^{\intercal}}{\sqrt{C}}\right) \mathcal{Q}_v^{\text{SA}} \right).
    \end{cases}\\
     &
    \begin{cases}
        \mathcal{Q}_q^{\text{CA}} = \operatorname{FC}_q^{\text{CA}}(\mathcal{Q}^{\text{SA}}),  \hat{\mathcal{T}}_k = \operatorname{FC}_k^{\text{CA}}(\hat{\mathcal{T}}),  \hat{\mathcal{T}}_v = \operatorname{FC}_v(\hat{\mathcal{T}}), \\ 
        \mathcal{Q}^{\text{CA}} = \operatorname{FC}_o^{\text{CA}}\left(\operatorname{softmax}\left(\frac{\mathcal{Q}_q^{\text{CA}}(\hat{\mathcal{T}}_k)^{\intercal}}{\sqrt{C}}\right) \hat{\mathcal{T}}_v\right).
    \end{cases}\\
\end{aligned}
\]
\vspace{-3mm}
\begin{equation}
    \begin{aligned}
    &  \hat{\mathcal{Q}} = \operatorname{MLP}(\operatorname{LayerNorm}(\mathcal{Q}^{\text{CA}}) + \mathcal{Q}), \quad \\
    &  B = \mathcal{H}_{\text{span}}(\hat{\mathcal{Q}}), \quad B = \{b_1, \cdots, b_M\}, \quad \\
\end{aligned}
\end{equation}
where $\operatorname{FC}(\cdot)$ denotes the fully connected projection layer, with $\text{SA}$ and $\text{CA}$ representing self-attention and cross-attention, respectively. Subscripts \(q\), \(k\), and \(v\) indicate query, key, and value computations. \(\operatorname{MLP}(\cdot)\) represents a nonlinear mapping with an activation function. \todo{\(\mathcal{H}_{\text{span}}(\cdot)\)} denotes the span prediction head, which takes the updated query features \(\hat{\mathcal{Q}}\) to predict the span set \(B\), where each \(b_i \in \mathbb{R}^2\) represents the (start, end) of a predicted span. 

\subsection{Query Grouping and Ranking}
\label{sec:4.2} 

In this section, we address the query conflicts identified in Sec~\ref{sec:3.2}: (1) the inability to distinguish between intra- and inter-segment queries (Figure~\ref{fig:3}), leading to query oscillation between segments without stable correspondence \textcolor{black}{(Line-266)}, and (2) inconsistencies in segment-query matching across layers, where different queries may be matched to ground-truth (GT) segments at each layer for optimization, hindering effective learning (Figure~\ref{fig:4}). 

To differentiate queries across segments, the challenge arises from GT segments that inherently share identical linguistic semantics, leading to highly similar queries. Therefore, a natural solution is to directly group queries, assigning each group to a specific GT segment for training and prediction. Instead of hard grouping with predefined rules or extra hyperparameters, we learn a soft grouping based on predicted temporal spans of queries, where closer spans indicate potential corresponding for the same segment. Specifically, for any two queries \(q_i\) and \(q_j\), we compute their span distance, which is then used in \textcolor{black}{Eq.~\ref{eq:4}} to learn their grouping:  
\vspace{-4mm}
\begin{equation}
\mathcal{S}^{\text{intra}}_{i,j} = \lVert b_i - b_j \rVert_2,
\end{equation}
where \(\lVert \cdot \rVert_2\) denotes the L2 norm. Notably, we use L2 distance instead of the commonly used L1 because, when two spans are close (normalized distances $\leq 1$), L2 distance decreases more sharply than L1, imposing a smaller penalty on minor differences \textcolor{black}{(see Appendix~\textcolor{iccvblue}{G})}. 

\begin{figure}[t]
    \centering
    \vspace{2mm}
    \includegraphics[width=0.9\linewidth]{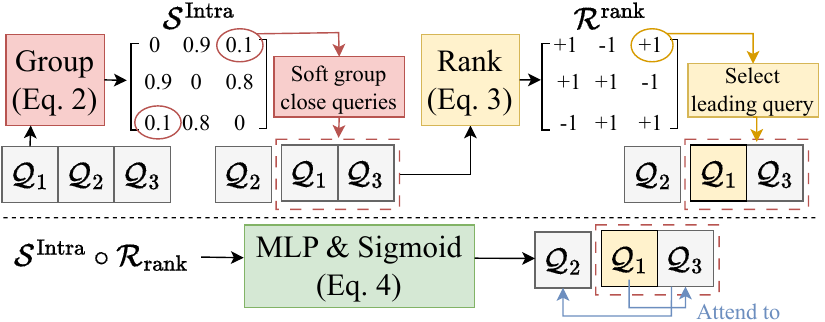}
    \vspace{-3mm}
    \caption{Overview of our Query Grouping and Ranking.}
    \label{fig:rebu}
    \vspace{-5mm}
\end{figure}

Meanwhile, to enhance cross-layer matching consistency, we draw inspiration from EASE-DETR~\cite{ease-detr}'s query ranking strategy in object detection. By ranking queries within a segment at each layer, the leading query is prioritized for segment matching in subsequent layers. However, directly ranking queries by prediction scores, as in EASE-DETR, proves ineffective \textcolor{black}{(see Appendix~\textcolor{iccvblue}{D})}. As analyzed in Sec.~\ref{sec:3.3}, high prediction/confidence scores do not guarantee precise predictions and may correspond to poor localization. Instead, we introduce an IoU head to assess localization accuracy and rank queries based on both factors, as follows:
\vspace{-3mm}
\begin{equation}
\label{eq:3}
R_{\text{rank}}(q_i,q_j)=\begin{cases} 
+1 & \mathcal{P}_i^{\text{cls}} \circ \mathcal{P}_i^{\text{IoU}} \geq \mathcal{P}_j^{\text{cls}} \circ \mathcal{P}_j^{\text{IoU}} \\
-1 & \mathcal{P}_i^{\text{cls}} \circ \mathcal{P}_i^{\text{IoU}} < \mathcal{P}_j^{\text{cls}} \circ \mathcal{P}_j^{\text{IoU}}
\end{cases}, 
\end{equation}
where \(\mathcal{P}^{\text{cls}}\) and \(\mathcal{P}^{\text{IoU}}\) denote the query's confidence score and IoU prediction, respectively, with subscripts \(i\) and \(j\) indicating queries \(q_i\) and \(q_j\). \(R_{\text{rank}}(q_i, q_j) = 1\) indicates that query \(q_i\) has a higher priority than query \(q_j\), considering both global alignment and local localization.

Finally, we incorporate query grouping within segments and intra-segment ranking to infer relative relationships, then reshape query-to-query self-attention \textcolor{black}{following~\cite{ease-detr}} to mitigate conflicts, as follows: 
\begin{equation}
\label{eq:4}
\begin{aligned}
    &\mathcal{S}^{\text{attn}} = \text{sigmoid}\left(\operatorname{MLP}(\mathcal{S}^{\text{intra}} \circ \mathcal{R}_{\text{rank}}\right), \\
    \mathcal{Q}^{\text{SA}} &= \operatorname{FC}_o^{\text{SA}}\left(\operatorname{softmax}\left(\left(\frac{\mathcal{Q}_q^{\text{SA}}(\mathcal{Q}_k^{\text{SA}})^{\intercal}}{\sqrt{C}}\right)\hspace{-2pt}\textbf{\colorbox{pink!50}{$\mathcal{S}^{\text{attn}}$}}\hspace{-2pt}\right) \mathcal{Q}_v^{\text{SA}}\right)
\end{aligned}
\end{equation}
where \(\mathcal{S}^{\text{attn}}\) represents query relationships, with higher values promoting high-quality queries to dominate predictions by integrating similar queries within the same group. 
 
\subsection{Global-Local Bridging}
\label{sec:4.3}
In this section, we address the query conflicts identified in Sec~\ref{sec:3.3}, specifically the conflict between a query's global semantic matching and local localization. 
Since queries overly rely on global semantics while neglecting local localization, our core idea is to reinforce fine-grained frame-level localization within the segment. 
Specifically, we introduce a query-to-frame matching mechanism and loss term to enhance localization beyond segment boundary prediction, ensuring alignment with in-segment frames while avoiding matches with those from other segments. The full sequence of frames within the segment, from start to finish, serves as a bridge linking global semantics to local boundaries. 

\begin{table*}
\centering
\small
\setlength\tabcolsep{2.4mm}
\renewcommand\arraystretch{0.8}
\begin{tabular}{lcccccccccc}
\toprule
\multirow{4}{*}{Method} & \multicolumn{5}{c}{test} & \multicolumn{5}{c}{val} \\

\cmidrule(rl){2-6} \cmidrule(rl){7-11}

& \multicolumn{2}{c}{R1} & \multicolumn{3}{c}{mAP} & \multicolumn{2}{c}{R1} & \multicolumn{3}{c}{mAP} \\

\cmidrule(rl){2-3} \cmidrule(rl){4-6} \cmidrule(rl){7-8} \cmidrule(rl){9-11}

& @0.5 & @0.7 & @0.5 & @0.75 & \multicolumn{1}{c}{Avg.} & @0.5 & @0.7 & @0.5 & @0.75 & Avg. \\%

\midrule

M-DETR~\cite{lei2021qvhighlights}{\scriptsize \textit{NeurIPS'21}} & 52.89 & 33.02 & 54.82 & 29.40 & 30.73 & 53.94 & 34.84 & - & - & 32.20 \\
UMT~\cite{liu2022umt}{\scriptsize \textit{CVPR'22}} & 56.23 & 41.18 & 53.83 & 37.01 & 36.12 & 60.26 & 44.26 & - & - & 38.59 \\
QD-DETR~\cite{moon2023qd-detr}{\scriptsize \textit{CVPR'23}} & 62.40 & 44.98 & 62.52 & 39.88 & 39.86 & 62.68 & 46.66 & 62.23 & 41.82 & 41.22 \\
UniVTG~\cite{lin2023univtg}{\scriptsize \textit{ICCV'23}} & 58.86 & 40.86 & 57.60 &  35.59 & 35.47 & 59.74 & - & - & - & 36.13 \\
EaTR~\cite{jang2023eatr}{\scriptsize \textit{ICCV'23}} & - & - & - & - & - & 61.36 & 45.79 & 61.86 & 41.91 & 41.74 \\
MomentDiff~\cite{li2023momentdiff}{\scriptsize \textit{NeurIPS'23}} & 57.42 & 39.66 & 54.02 & 35.73 & 35.95 & - & - & - & - & - \\
TR-DETR~\cite{sun2024trdetr}{\scriptsize \textit{AAAI'24}} & 64.66 & 48.96 & 63.98 & 43.73 & 42.62 & 67.10 & 51.48 & \underline{}{66.27} & 46.42 & 45.09 \\
UVCOM~\cite{xiao2024uvcom}{\scriptsize \textit{CVPR'24}} & 63.55 & 47.47 & 63.37 & 42.67 & 43.18 & 65.10 & 51.81 & - & - & 45.79 \\
TaskWeave~\cite{yang2024taskweave}{\scriptsize \textit{CVPR'24}} & - & - & - & - & - & 64.26 & 50.06 & 65.39 & 46.47 & 45.38 \\
BAM-DETR~\cite{lee2025bamdetr}{\scriptsize \textit{ECCV‘24}} & 62.71 & 48.64 & \underline{}{64.57} & \underline{}{46.33} & \underline{}{45.36} & 65.10 & 51.61 & 65.41 & \underline{}{48.56} & \underline{}{47.61}\\
CG-DETR~\cite{moon2023cgdetr}{\scriptsize \textit{Arxiv‘24}} & \underline{}{65.43} & 48.38 & 64.51 & 42.77 & 42.86 & \underline{}{67.35} & \underline{}{52.06} & 65.57 & 45.73 & 44.93 \\
SpikeMba~\cite{li2024spikemba}{\scriptsize \textit{Arxiv‘24}} & 64.13 & \underline{}{49.42} & - & 43.67 & 43.79 & 65.32 & 51.33 & - & 44.96 & 44.84\\
\midrule
\textbf{Sim-DETR(Ours)} & \textbf{67.64} & \textbf{50.91} & \textbf{67.81} & \textbf{47.59} & \textbf{46.93} & \textbf{69.48} & \textbf{54.06} & \textbf{69.70} & \textbf{51.11} & \textbf{49.50} \\

\bottomrule
\end{tabular}
\vspace{-5pt}
\caption{Experimental results on the QVHighlights benchmark, comparing ours with state-of-the-art methods.}
\label{tab:1}
\vspace{-4pt}
\end{table*}
\begin{table*}
    \small
    \centering
    \setlength\tabcolsep{3mm}
    \renewcommand\arraystretch{0.78}
    \begin{tabular}{lcccccccc}
    \toprule
    \multirow{2.5}{*}{Method} & \multicolumn{4}{c}{TACoS} & \multicolumn{4}{c}{Charades-STA}\\
    \cmidrule(rl){2-5} \cmidrule(rl){6-9}
    & R1@0.3 & R1@0.5 & R1@0.7 & mIoU & R1@0.3 & R1@0.5 & R1@0.7 & mIoU\\
    \midrule
    2D-TAN~\cite{zhang2020tan} & 40.01 & 27.99 & 12.92 & 27.22 & 58.76 & 46.02 & 27.50 & 41.25 \\
    M-DETR~\cite{lei2021qvhighlights} & 37.97 & 24.67 & 11.97 & 25.49 & 65.83 & 52.07 & 30.59 & 45.54 \\
    MomentDiff~\cite{li2023momentdiff} & 44.78 & 33.68 & - & - & - & 55.57 & 32.42 & - \\
    QD-DETR~\cite{moon2023qd-detr} & - & - & - & - & - & 57.31 & 32.55 & - \\
    UniVTG~\cite{lin2023univtg} & 51.44 & 34.97 & 17.35 & 33.60 & 70.81 & 58.01 & 35.65 & 50.10 \\
    CG-DETR~\cite{moon2023cgdetr} & 52.23 & 39.61 & 22.23 & 36.48 & 70.43 & 58.44 & 36.34 & 50.13 \\
    UVCOM~\cite{xiao2024uvcom} & - & 36.39 & 23.32 & - & - & 59.25 & 36.64 & - \\
    SpikeMba~\cite{li2024spikemba} & 51.98 & 39.34 & 22.83 & 35.81 & 71.24 & 59.65 & 36.12 & 51.74 \\
    \midrule
    \textbf{Sim-DETR(Ours)} & \textbf{57.06} & \textbf{42.79} & \textbf{26.82} & \textbf{39.44} & \textbf{73.09} & \textbf{61.34} & \textbf{39.62} & \textbf{52.56} \\
    \bottomrule
    \end{tabular}
\vspace{-2mm}
\caption{Comparison results on the TACoS and Charades-STA benchmarks.}
\label{tab:2}
\vspace{-4mm}
\end{table*}

During training, we compute the semantic similarity between each query \( q_i \) and all frames within its corresponding ground-truth span, maximizing similarity to ensure full alignment with in-segment frames while minimizing similarity to out-segment frames.
The computation process is outlined as follows, 
\vspace{-3mm}
\begin{equation}
\begin{aligned}
&z = \operatorname{sigmoid}(\tau \cdot \cos(q_i, \hat{\mathcal{T}})),\\
\mathcal{L}_{\text{bridge}} = &\lambda_{\text{bridge}}\frac{-\sum_{j} z_j \mathbb{I}[b^{\text{gt}}_i]_j} {\sum_{j} z_j (1-\mathbb{I}[b^{\text{gt}}_i]_j) + \sum_{j} \mathbb{I}[b^{\text{gt}}_i]_j},
\end{aligned}
\end{equation}
where \(z \in \mathbb{R}^N\) donates query-to-frame similarity, calculated using the cosine similarity between the query $q_i$ and the $N$ frame features \(\hat{\mathcal{T}}\) defined in \textcolor{black}{Line-347}, and $\tau$ is a learnable scaling coefficient. \( b_i^{\text{gt}} \) is the ground-truth span corresponding to query \( q_i \), while \( \mathbb{I}[b_i^{\text{gt}}] \in \mathbb{R}^{N} \) is an indicator function that assigns 1 to frames within the ground truth and 0 otherwise, with \( j \) representing the frame index. Thus, the numerator in \(\mathcal{L}_{\text{bridge}}\), \ie, \(-\sum_{j} z_j \mathbb{I}[b^{\text{gt}}_i]_j\), encourages the query to be similar to every frame within its corresponding segment. Meanwhile, the denominator's first term, \(\sum_{j} z_j (1-\mathbb{I}[b^{\text{gt}}_i]_j)\), represents query \( q_i \)'s similarity to out-segment frames, promoting its minimization. The denominator's second term, \(\sum_{j} \mathbb{I}[b^{\text{gt}}_i]_j\), represents the length of the ground-truth segment, serving as a normalization factor. And \(\lambda_{\text{bridge}}\) is a hyperparameter that controls the loss weight. 

\subsection{Overall Loss}
\label{sec:4.4} 
The overall loss is:
\(
    \mathcal{L} = \mathcal{L}_{\text{MD}} + \lambda_{\text{bridge}}\mathcal{L}_{\text{bridge}} + \lambda_{\text{iou}}\mathcal{L}_{\text{iou}},
\)
where \( \mathcal{L}_{\text{MD}} \) includes the L1, gIoU, classification, and saliency losses, as in Moment DETR~\cite{lei2021detecting}. 
\(\mathcal{L}_{\text{iou}}\) is the loss for the IoU head introduced in \textcolor{black}{Line-406}, designed to assist in ranking queries using \( \mathcal{P}^{\text{IoU}}_i \) and \( \mathcal{P}^{\text{IoU}}_j \) in \textcolor{black}{Eq~\ref{eq:3}}. The IoU head implementation follows~\cite{lee2025bamdetr,sun2024rgtr}. 
Ablations for hyperparameters \( \lambda_{\text{bridge}} \) and \( \lambda_{\text{iou}} \) are \textcolor{black}{detailed in Appendix~\textcolor{cvprblue}{E}.}
\section{Experiments}
\subsection{Datasets and Implementation Details}
\noindent \textbf{Datasets.} Following~\cite{lei2021qvhighlights,moon2023qd-detr,lin2023univtg,xiao2024uvcom, li2024spikemba}, we conduct comprehensive evaluation experiments across three datasets: QVHighlights~\cite{lei2021qvhighlights}, Charades-STA~\cite{gao2017ctrl}, and TACoS~\cite{regneri2013tacos}. The QVHighlights dataset contains 10,310 text expressions and 18,367 distinct events, with most expressions corresponding to multiple events. 
Charades-STA is derived from a collection of 9,848 indoor scene videos, encompassing 16,128 events.TACoS includes only 273 videos, with an average length approaching five minutes per video. 

For a fair comparison, we adopt the same evaluation metrics as in previous methods. On the QVHighlights benchmark, we use Recall and Mean Average Precision (mAP) as primary metrics, while for Charades-STA and TACoS, we employ Recall and Mean Intersection over Union (mIoU).

\noindent \textbf{Implementation Details.} We set the input video resolution to 224$\times$224 and represent each video by concatenating the CLIP~\cite{radford2021clip} [CLS] tokens with SlowFast~\cite{feichtenhofer2019slowfast} features. For the Charades-STA and TACoS datasets, we additionally constructed two experiments using I3D~\cite{carreira2017i3d} and VGG~\cite{simonyan2014vgg} as visual encoders, with Glove~\cite{pennington2014glove} embeddings for text features. The model is trained for 200 epochs on a single NVIDIA A40 GPU, using the AdamW~\cite{loshchilov2017adamw} optimizer with an initial learning rate of 1e-4. \note{Note that, for a fair comparison, the total number of layers in our Sim-DETR is set to 6, the same as recent works~\cite{moon2023cgdetr, jang2023eatr,xiao2024uvcom}}.
\begin{table}
    \centering
    \small
\setlength\tabcolsep{3.5mm}
\renewcommand\arraystretch{0.8}
\vspace{4pt}
    \begin{tabular}{llcc}
    \toprule
    Backbone & Method & R@0.5 & R@0.7\\
    \midrule
    \multirow{9}{*}{VGG~\cite{simonyan2014vgg}} 
    & 2D-TAN~\cite{zhang2020tan} & 40.94 & 22.85 \\
    & FVMR~\cite{gao2021fvmr} & 42.36 & 24.14 \\
    & SSRN~\cite{zhu2023ssrn} & 46.72 & 27.98 \\
    & UMT~\cite{liu2022umt} & 48.31 & 29.25 \\
    & MomentDiff~\cite{li2023momentdiff} & 51.94 & 28.25 \\
    & QD-DETR~\cite{moon2023qd-detr} & 52.77 & 31.13 \\
    & TR-DETR~\cite{sun2024trdetr} & 53.47 & 30.81 \\
    & CG-DETR~\cite{moon2023cgdetr} & 55.22 & 34.19 \\
    \rowcolor{pink!30}
    & \textbf{Sim-DETR(Ours)} & \textbf{55.97} & \textbf{35.38} \\
    \midrule
    \multirow{6}{*}{I3D~\cite{carreira2017i3d}}
    & MAN~\cite{zhang2019man} & 46.53 & 22.72 \\
    & VSLNet~\cite{zhang2020vslnet} & 47.31 & 30.19 \\
    & QD-DETR~\cite{moon2023qd-detr} & 50.67 & 31.02 \\
    & TR-DETR~\cite{sun2024trdetr} & 55.51 & 33.66 \\
    & TaskWeave~\cite{yang2024taskweave} & 53.36 & 31.40 \\
    \rowcolor{pink!30}
    & \textbf{Sim-DETR(Ours)} & \textbf{57.55} & \textbf{35.73} \\
    \bottomrule
    \end{tabular}
    \vspace{-2mm}
    \caption{Experimental results with VGG or I3D as backbone.}
    \label{tab:3}
    \vspace{-6mm}
\end{table} 
\subsection{Comparison with State-of-the-Art Methods}
As shown in Tables~\ref{tab:1}\textcolor{cvprblue}{-}\ref{tab:3}, we report detailed evaluation results across the QVHighlights, Charades-STA, and TACoS benchmarks. Our Sim-DETR consistently outperforms all state-of-the-art methods (SOTAs) across all metrics.

\noindent \textbf{Results on QVHighlights} are shown in Table~\ref{tab:1}.
Compared to the latest published work, BAM-DETR~\cite{lee2025bamdetr}, our Sim-DETR outperforms it by an average of \todo{+2.88\%} across all metrics. More importantly, it achieves remarkable gains of \todo{+4.93\%} and \todo{+2.27\%} in R1@0.5 and R1@0.7, respectively.
In comparison TR-DETR~\cite{sun2024trdetr}, which uses standard DETR~\cite{carion2020detr} as its basic framework and is used as our baseline model, our Sim-DETR achieves an average improvement of \todo{+4.31\%} and \todo{+4.41\%} in mAP for test and validation, respectively. These improvements underscore the effectiveness of our Sim-DETR in fundamentally addressing conflicts within and between queries. We further conduct a comparison with SpikeMba~\cite{li2024spikemba}, a contemporary method built on the Mamba~\cite{gu2023mamba} framework. Without any complex module designs, the consistent improvements across all metrics highlight the simplicity and effectiveness of our Sim-DETR. Sim-DETR establishes an efficient baseline for the task, significantly advancing research in TSG. 

\noindent \textbf{Results on Charades-STA and TACoS} are presented in Table~\ref{tab:2}, where we consistently outperform all methods on both datasets. Compared to the state-of-the-art methods, we achieve a \todo{+1.97\%} improvement on the Charades-STA dataset (with SpikeMba as the SOTA) and a \todo{+3.89\%} improvement on the TACoS dataset (with CG-DETR~\cite{moon2023cgdetr} as the SOTA). These enhancements further demonstrate the outstanding capability of our method in temporal localization. More importantly, they confirm that our Sim-DETR exhibits significant performance advantages across various scenarios, types, and lengths of videos, highlighting its robustness and generalizability.
\begin{table}
\small
\begin{minipage}{0.45\textwidth}
\centering
\setlength\tabcolsep{1.7mm}
\renewcommand\arraystretch{0.8}
\begin{tabular}{lcccccc}
\toprule
\multirow{2.5}{*}{Method} & \multicolumn{2}{c}{R1} & & \multicolumn{3}{c}{mAP} \\
\cmidrule{2-3} \cmidrule{5-7}
& @0.5 & @0.7 & & @0.5 & @0.75 & Avg. \\
\midrule
Baseline & 65.48 & 50.84 & & 65.34 & 45.82 & 44.97 \\
+ $\mathcal{L}_{\text{iou}}$ & 66.58 & 51.94 & & 65.68 & 46.30 & 45.22 \\
+ QGR & 68.77 & 52.26 & & 67.72 & 48.28 & 47.03 \\
+ GLB & 67.16 & 52.77 & & 68.58 & 48.88 & 48.17 \\
+ QGR,GLB & \textbf{69.48} & \textbf{54.06} & & \textbf{69.70} & \textbf{51.11} & \textbf{49.50} \\
\bottomrule
\end{tabular}
\vspace{-5pt}
\caption{Ablation study on components.}
\vspace{5pt}
\label{tab:4}
\end{minipage}

\begin{minipage}{0.45\textwidth}
\centering
\setlength\tabcolsep{1mm}
\renewcommand\arraystretch{0.8}
\begin{tabular}{lc|ccc}
\toprule
 & Inner Relevance & Outer Global & Outer Local & \makecell{mAP}\\
\midrule
(1) & span border dist & confidence pred & IoU pred & \textbf{49.50}\\
\midrule
(2) & \cellcolor{pink!30} center dist &  &  & 48.59\\
(3) & \cellcolor{pink!30} span IoU & \multirow{-2}{*}{confidence pred} & \multirow{-2}{*}{IoU pred} & 48.94\\
\midrule
(4) & \multirow{-1}{*}{ span border dist} &\cellcolor{pink!30} w/o & \multirow{-1}{*}{IoU pred} & 48.93\\
\midrule
(5) &  &  & \cellcolor{pink!30} w/o & 48.66\\
(6) & \multirow{-2}{*}{ span border dist} & \multirow{-3}{*}{confidence pred} & \cellcolor{pink!30} center pred & 48.97\\
\bottomrule
\end{tabular}
\vspace{-5pt}
\caption{Impact of different measures.}
\label{tab:5}
\vspace{-6mm}
\end{minipage}
\end{table}

\noindent \textbf{Results with Different Backbones.} 
Following previous works, we conduct two additional experiments on the QVHighlights dataset. These experiments use VGG and I3D for visual feature extraction and use Glove~\cite{pennington2014glove}embedding as text features. Table~\ref{tab:3} presents the detailed results of these two experiments, where we achieve state-of-the-art results in both cases. Notably, for the video feature extractor I3D, we surpass the latest published method by \todo{+4.26\%}. These results further demonstrate the excellent generalization capabilities of our Sim-DETR across different backbones.

\subsection{Ablation Study}
\noindent \textbf{Roadmap} for building a simple yet effective TSG baseline is presented in Table~\ref{tab:4}. \note{We employ TR-DETR without MR2HD module as our baseline, which achieves an mAP of 44.97\%.} Since our Query Grouping and Ranking (QGR) utilizes $\mathcal{L}_{\text{iou}}$ proposed by \cite{lee2025bamdetr, sun2024rgtr}, we additionally conduct an ablation study to isolate the impact of $\mathcal{L}_{\text{iou}}$. (1) \todo{QGR}: To mitigate conflicts and optimization challenges arising from overly similar semantics between queries, we introduce the QGR, which improves the baseline by +2.07\% on mAP, representing a subtle yet significant enhancement. (2) \todo{Global-Local Bridging (GLB)}: To address conflicts and ambiguities between global semantics and local localization within queries, we introduce GLB to enable seamless transformation through query-to-frame alignment. Experimental results show a performance boost to 48.17\% on mAP, indicating the effectiveness of GLB in bridging global-to-local transformations. However, without QGR to address semantic conflicts in temporal DETR, the recall predictably decreased by -3.1\%. (3) Overall Framework: With these adjustments, our Sim-DETR achieves a remarkable SOTA performance: 49.50\% on mAP and 69.48\% on R1@0.5.

\noindent \textbf{The Effects of Different Conflict Metrics in QGR.} Table \ref{tab:5} summarizes five experiments on conflict metrics in QGR. (1) The first row displays our final implementation, achieving 49.50\% mAP. (2) \& (3) We evaluate L2 center distance and span IoU for inter-query relationships. Center distance alone yields 48.59\% mAP, highlighting its inadequacy in segment representation. Span IoU outperforms center distance but lags by 0.56\% compared to boundary distance, likely due to IoU’s broader tolerance, which reduces precision in query relevance. (4) To assess outer global semantic alignment, we remove global semantics, resulting in a -0.57\% drop, emphasizing their role in query interaction and quality. (5) Finally, we test local location metrics in query prioritization. Excluding the outer local metric significantly lowers performance (-0.84\% mAP), indicating that without local information, queries encounter conflicts due to overly generalized global semantics, leading to random matches. Consistent with (2), center localization accuracy shows similar limitations, with a 0.53\% decrease.
\section{Conclusion}
In this paper, we propose a concise and efficient temporal sentence grounding framework Sim-DETR, introducing two simple modifications to the decoder for resolving conflicts both between and within queries. Experimental results demonstrate that our approach significantly outperforms state-of-the-art methods across all benchmarks. 

\noindent\textbf{Acknowledgment:} This work is supported by the National Natural Science Foundation of China (No.62206174). 
{
    \small
    \bibliographystyle{ieeenat_fullname}
    \bibliography{main}
}

\end{document}